%% file: sgparser.tex
\documentclass[11pt,a4paper]{article}
\usepackage[hyperref]{naaclhlt2018}
\usepackage{times}
\usepackage{latexsym}
\usepackage{amsmath}
\usepackage{amssymb}
\usepackage{booktabs}
\usepackage{graphicx}
\usepackage{multirow}
\usepackage[ruled, vlined, linesnumbered]{algorithm2e}
\SetArgSty{textup}

\usepackage[misc]{ifsym}

\usepackage[english]{babel}
\usepackage[autostyle]{csquotes}
\MakeOuterQuote{"}

\usepackage{url}

\aclfinalcopy 

\title{Scene Graph Parsing as Dependency Parsing}

\author{Yu-Siang Wang \\
  National Taiwan University \\
  {\tt b03202047@ntu.edu.tw} \\\And
  Chenxi Liu$^{(\textrm{\Letter})}$ \\
  Johns Hopkins University \\
  {\tt cxliu@jhu.edu} \\\AND
  Xiaohui Zeng \\
  Hong Kong University of Science and Technology \\
  {\tt xzengaf@connect.ust.hk} \\\And
  Alan Yuille \\
  Johns Hopkins University \\
  {\tt alan.yuille@jhu.edu}  
  }
\date{}

\begin{document}
\maketitle
\begin{abstract}
In this paper, we study the problem of parsing structured knowledge graphs from textual descriptions.
In particular, we consider the scene graph representation \citep{DBLP:conf/cvpr/JohnsonKSLSBL15} that considers objects together with their attributes and relations: this representation has been proved useful across a variety of vision and language applications.
We begin by introducing an alternative but equivalent edge-centric view of scene graphs that connect to dependency parses.
Together with a careful redesign of label and action space, we combine the two-stage pipeline used in prior work (generic dependency parsing followed by simple post-processing) into one, enabling end-to-end training.
The scene graphs generated by our learned neural dependency parser achieve an F-score similarity of 49.67\% to ground truth graphs on our evaluation set, surpassing best previous approaches by 5\%.
We further demonstrate the effectiveness of our learned parser on image retrieval applications.\footnote{Code is available at \url{https://github.com/Yusics/bist-parser/tree/sgparser}}
\end{abstract}

\input{intro}
\input{related}
\input{task}
\input{parsing}
\input{exp}
\input{conc}

\section*{Acknowledgments}

The majority of this work was done when YSW and XZ were visiting Johns Hopkins University.
We thank Peter Anderson, Sebastian Schuster, Ranjay Krishna, Tsung-Yi Lin for comments and help regarding the experiments.
We also thank Tianze Shi, Dingquan Wang, Chu-Cheng Lin for discussion and feedback on the draft.
This work was sponsored by the National Science Foundation Center for Brains, Minds, and Machines NSF CCF-1231216.
CL also acknowledges an award from Snap Inc.

\bibliography{sgparser}
\bibliographystyle{acl_natbib}

\end{document}

%% file: intro.tex
\section{Introduction}

Recent years have witnessed the rise of interest in many tasks at the intersection of computer vision and natural language processing, including semantic image retrieval \citep{DBLP:conf/cvpr/JohnsonKSLSBL15, DBLP:journals/corr/VendrovKFU15}, image captioning \citep{DBLP:journals/corr/MaoXYWY14a,DBLP:conf/cvpr/KarpathyL15,DBLP:conf/cvpr/DonahueHGRVDS15,DBLP:conf/aaai/LiuMSY17}, visual question answering \citep{DBLP:conf/iccv/AntolALMBZP15,DBLP:conf/cvpr/ZhuGBF16,DBLP:conf/cvpr/AndreasRDK16}, and referring expressions \citep{DBLP:conf/cvpr/HuXRFSD16,DBLP:conf/cvpr/MaoHTCY016,DBLP:conf/iccv/LiuLSYLY17}.
The pursuit for these tasks is in line with people's desire for high level understanding of visual content, in particular, using textual descriptions or questions to help understand or express images and scenes.

What is shared among all these tasks is the need for a \emph{common representation} to establish connection between the two different modalities.
The majority of recent works handle the vision side with convolutional neural networks, and the language side with recurrent neural networks \citep{DBLP:journals/neco/HochreiterS97,DBLP:conf/emnlp/ChoMGBBSB14} or word embeddings \citep{DBLP:journals/corr/abs-1301-3781,DBLP:conf/emnlp/PenningtonSM14}.
In either case, neural networks map original sources into a semantically meaningful \citep{DBLP:conf/icml/DonahueJVHZTD14,DBLP:journals/corr/abs-1301-3781} vector representation that can be aligned through end-to-end training \citep{DBLP:conf/nips/FromeCSBDRM13}. 
This suggests that the vector embedding space is an appropriate choice as the common representation connecting different modalities (see e.g. \citet{DBLP:journals/corr/KaiserGSVPJU17}).

\begin{figure}[t]
\centering
\includegraphics[width=\linewidth]{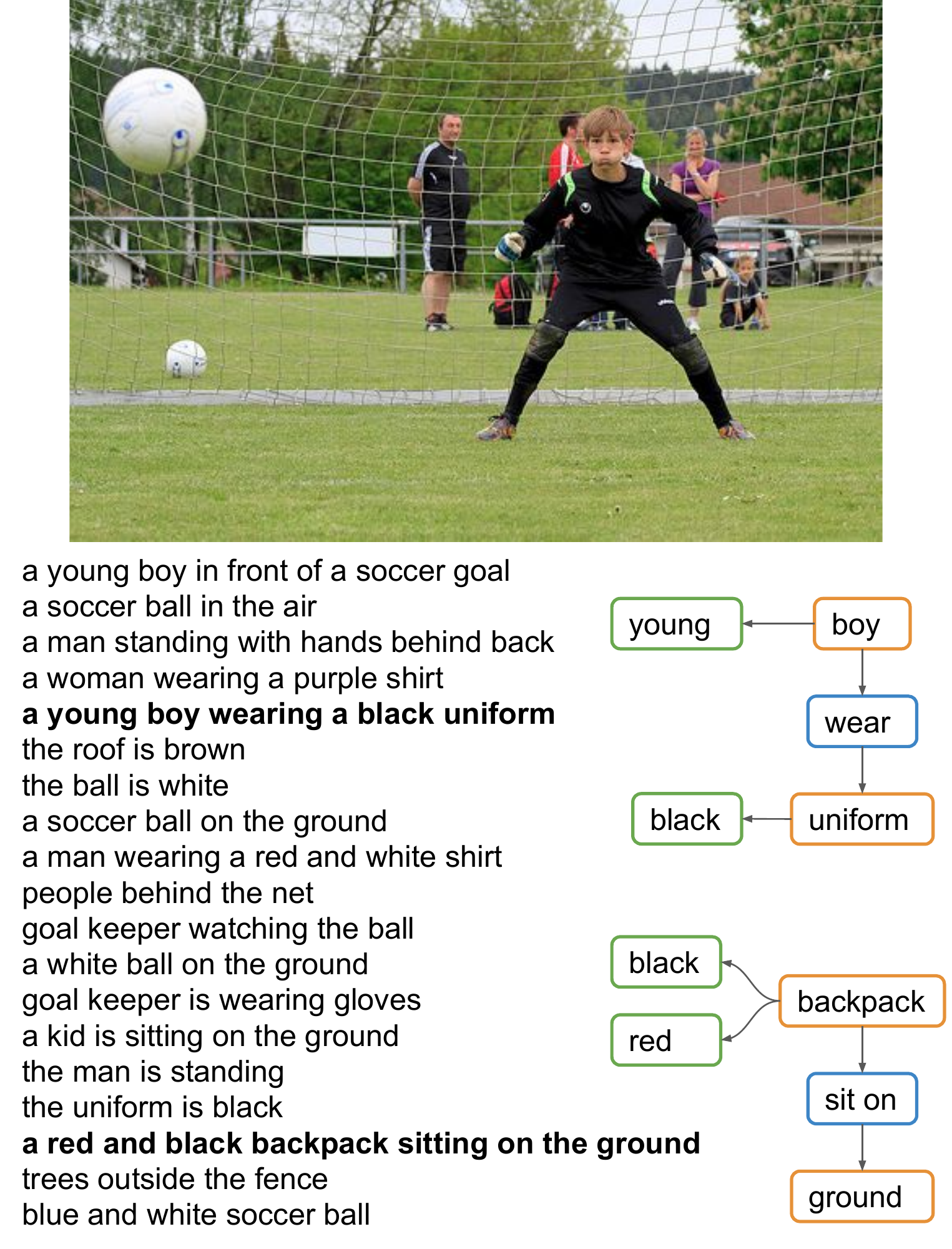}
\caption{Each image in the Visual Genome \citep{DBLP:journals/ijcv/KrishnaZGJHKCKL17} dataset contains tens of region descriptions and the region scene graphs associated with them.
In this paper, we study how to generate high quality scene graphs (two such examples are shown in the figure) from textual descriptions, without using image information.}
\label{fig: fig1}
\end{figure}

While the dense vector representation yields impressive performance, it has an unfortunate limitation of being less intuitive and hard to interpret. 
Scene graphs \citep{DBLP:conf/cvpr/JohnsonKSLSBL15}, on the other hand, proposed a type of directed graph to encode information in terms of objects, attributes of objects, and relationships between objects (see Figure~\ref{fig: fig1} for visualization).
This is a more structured and explainable way of expressing the knowledge from either modality, and is able to serve as an alternative form of common representation.
In fact, the value of scene graph representation has already been proven in a wide range of visual tasks, including semantic image retrieval \citep{DBLP:conf/cvpr/JohnsonKSLSBL15}, caption quality evaluation \citep{DBLP:conf/eccv/AndersonFJG16}, etc.
In this paper, we focus on scene graph generation from textual descriptions.

Previous attempts at this problem \citep{schuster2015generating,DBLP:conf/eccv/AndersonFJG16} follow the same spirit.
They first use a dependency parser to obtain the dependency relationship for all words in a sentence, and then use either a rule-based or a learned classifier as post-processing to generate the scene graph.
However, the rule-based classifier cannot learn from data, and the learned classifier is rather simple with hand-engineered features.
In addition, the dependency parser was trained on linguistics data to produce complete dependency trees, some parts of which may be redundant and hence confuse the scene graph generation process.

Therefore, our model abandons the two-stage pipeline, and uses a single, customized dependency parser instead.
The customization is necessary for two reasons. 
First is the difference in label space. 
Standard dependency parsing has tens of edge labels to represent rich relationships between words in a sentence, but in scene graphs we are only interested in three types, namely objects, attributes, and relations.
Second is whether every word needs a head. 
In some sense, the scene graph represents the "skeleton" of the sentence, which suggests that empty words are unlikely to be included in the scene graph.
We argue that in scene graph generation, it is unnecessary to require a parent word for every single word.

We build our model on top of a neural dependency parser implementation \citep{DBLP:journals/tacl/KiperwasserG16} that is among the state-of-the-art.
We show that our carefully customized dependency parser is able to generate high quality scene graphs by learning from data.
Specifically, we use the Visual Genome dataset \citep{DBLP:journals/ijcv/KrishnaZGJHKCKL17}, which provides rich amounts of region description - region graph pairs.
We first align nodes in region graphs with words in the region descriptions using simple rules, and then use this alignment to train our customized dependency parser.
We evaluate our parser by computing the F-score between the parsed scene graphs and ground truth scene graphs.
We also apply our approach to image retrieval to show its effectiveness.

%% file: related.tex
\section{Related Works}

\subsection{Scene Graphs}

The scene graph representation was proposed in \citet{DBLP:conf/cvpr/JohnsonKSLSBL15} as a way to represent the rich, structured knowledge within an image.
The nodes in a scene graph represent either an object, an attribute for an object, or a relationship between two objects.
The edges depict the connection and association between two nodes.
This representation is later adopted in the Visual Genome dataset \citep{DBLP:journals/ijcv/KrishnaZGJHKCKL17}, where a large number of scene graphs are annotated through crowd-sourcing.

The scene graph representation has been proved useful in various problems including semantic image retrieval \citep{DBLP:conf/cvpr/JohnsonKSLSBL15}, visual question answering \citep{DBLP:journals/corr/TeneyLH16}, 3D scene synthesis \citep{DBLP:conf/emnlp/ChangSM14}, and visual relationship detection \citep{DBLP:conf/eccv/LuKBL16}.
Excluding \citet{DBLP:conf/cvpr/JohnsonKSLSBL15} which used ground truth, scene graphs are obtained either from images \citep{DBLP:conf/cvpr/DaiZL17,DBLP:conf/cvpr/XuZCF17,DBLP:journals/corr/LiOZWW17} or from textual descriptions \citep{schuster2015generating,DBLP:conf/eccv/AndersonFJG16}.
In this paper we focus on the latter.

In particular, parsed scene graphs are used in \citet{schuster2015generating} for image retrieval. 
We show that with our more accurate scene graph parser, performance on this task can be further improved.

\subsection{Parsing to Graph Representations}

The goal of dependency parsing \citep{DBLP:series/synthesis/2009Kubler} is to assign a parent word to every word in a sentence, and every such connection is associated with a label.
Dependency parsing is particularly suitable for scene graph generation because it directly models the relationship between individual words without introducing extra nonterminals.
In fact, all previous work \citep{schuster2015generating,DBLP:conf/eccv/AndersonFJG16} on scene graph generation run dependency parsing on the textual description as a first step, followed by either heuristic rules or simple classifiers.
Instead of running two separate stages, our work proposed to use a single dependency parser that is end-to-end.
In other words, our customized dependency parser generates the scene graph in an online fashion as it reads the textual description once from left to right.

In recent years, dependency parsing with neural network features \citep{DBLP:conf/emnlp/ChenM14,DBLP:conf/acl/DyerBLMS15,DBLP:conf/acl/CrossH16,DBLP:journals/tacl/KiperwasserG16,DBLP:journals/corr/DozatM16,DBLP:conf/emnlp/ShiHL17} has shown impressive performance.
In particular, 
\citet{DBLP:journals/tacl/KiperwasserG16} used bidirectional LSTMs to generate features for individual words, which are then used to predict parsing actions.
We base our model on \citet{DBLP:journals/tacl/KiperwasserG16} for both its simplicity and good performance.

Apart from dependency parsing, Abstract Meaning Representation (AMR) parsing \citep{DBLP:conf/acl/FlaniganTCDS14,DBLP:conf/acl/WerlingAM15,DBLP:conf/naacl/WangXP15,DBLP:conf/acl/KonstasIYCZ17} may also benefit scene graph generation.
However, as first pointed out in \citet{DBLP:conf/eccv/AndersonFJG16}, the use of dependency trees still appears to be a common theme in the literature, and we leave the exploration of AMR parsing for scene graph generation as future work.

More broadly, our task also relates to entity and relation extraction, e.g. \citet{DBLP:conf/acl/KatiyarC17}, but there object attributes are not handled.
Neural module networks \citep{DBLP:conf/cvpr/AndreasRDK16} also use dependency parses, but they translate questions into a series of actions, whereas we parse descriptions into their graph form.
Finally, \citet{DBLP:journals/tacl/KrishnamurthyK13} connected parsing and grounding by training the parser in a weakly supervised fashion.

%% file: task.tex
\section{Task Description}

In this section, we begin by reviewing the scene graph representation, and show how its nodes and edges relate to the words and arcs in dependency parsing.
We then describe simple yet reliable rules to align nodes in scene graphs with words in textual descriptions, such that customized dependency parsing, described in the next section, may be trained and applied.

\subsection{Scene Graph Definition}

There are three types of nodes in a scene graph: object, attribute, and relation.
Let $\mathcal{O}$ be the set of object classes, $\mathcal{A}$ be the set of attribute types, and $\mathcal{R}$ be the set of relation types.
Given a sentence $s$, our goal in this paper is to parse $s$ into a scene graph:
\begin{equation}
G(s) = \langle O(s), A(s), R(s) \rangle
\end{equation}
where $O(s) = \{o_1(s), \hdots, o_m(s)\}, o_i(s) \in \mathcal{O}$ is the set of object instances mentioned in $s$, $A(s) \subseteq O(s) \times \mathcal{A}$ is the set of attributes associated with object instances, and $R(s) \subseteq O(s) \times \mathcal{R} \times O(s)$ is the set of relations between object instances.

$G(s)$ is a graph because we can first create an object node for every element in $O(s)$;
then for every $(o, a)$ pair in $A(s)$, we create an attribute node and add an unlabeled edge $o \rightarrow a$;
finally for every $(o_1, r, o_2)$ triplet in $R(s)$, we create a relation node and add two unlabeled edges $o_1 \rightarrow r$ and $r \rightarrow o_2$.
The resulting directed graph exactly encodes information in $G(s)$.
We call this the \textbf{node-centric} graph representation of a scene graph.

We realize that a scene graph can be equivalently represented by no longer distinguishing between the three types of nodes, yet assigning labels to the edges instead.
Concretely, this means there is now only one type of node, but we assign a \texttt{ATTR} label for every $o \rightarrow a$ edge, a \texttt{SUBJ} label for every $o_1 \rightarrow r$ edge, and a \texttt{OBJT} label for every $r \rightarrow o_2$ edge.
We call this the \textbf{edge-centric} graph representation of a scene graph.

We can now establish a connection between scene graphs and dependency trees.
Here we only consider scene graphs that are acyclic\footnote{In Visual Genome, only 4.8\% region graphs have cyclic structures.}.
The edge-centric view of a scene graph is very similar to a dependency tree: they are both directed acyclic graphs where the edges/arcs have labels.
The difference is that in a scene graph, the nodes are the objects/attributes/relations and the edges have label space $\{$\texttt{ATTR}, \texttt{SUBJ}, \texttt{OBJT}$\}$, whereas in a dependency tree, the nodes are individual words in a sentence and the edges have a much larger label space.


\subsection{Sentence-Graph Alignment}


We have shown the connection between nodes in scene graphs and words in dependency parsing.
With alignment between nodes in scene graphs and words in the textual description, scene graph generation and dependency parsing becomes equivalent: we can construct the generated scene graph from the set of labeled edges returned by the dependency parser.
Unfortunately, such alignment is not provided between the region graphs and region descriptions in the Visual Genome \citep{DBLP:journals/ijcv/KrishnaZGJHKCKL17} dataset.
Here we describe how we use simple yet reliable rules to do sentence-graph (word-node) alignment.

There are two strategies that we could use in deciding whether to align a scene graph node $d$ (whose label space is $\mathcal{O} \cup \mathcal{A} \cup \mathcal{R}$) with a word/phrase $w$ in the sentence:
\begin{itemize}
\item Word-by-word match (WBW): $d \leftrightarrow w$ only when $d$'s label and $w$ match word-for-word.
\item Synonym match (SYN)\footnote{This strategy is also used in \cite{DBLP:conf/wmt/DenkowskiL14} and \cite{DBLP:conf/eccv/AndersonFJG16}.}: $d \leftrightarrow w$ when the wordnet synonyms of $d$'s label contain $w$. 
\end{itemize}
Obviously WBW is a more conservative strategy than SYN.

We propose to use two cycles and each cycle further consists of three steps, where we try to align objects, attributes, relations in that order.
The pseudocode for the first cycle is in Algorithm~\ref{alg: align-first-cycle}.
The second cycle repeats line 4-15 immediately afterwards, except that in line 6 we also allow SYN.
Intuitively, in the first cycle we use a conservative strategy to find "safe" objects, and then scan for their attributes and relations.
In the second cycle we relax and allow synonyms in aligning object nodes, also followed by the alignment of attribute and relation nodes.

The ablation study of the alignment procedure is reported in the experimental section.

\begin{algorithm}[t]
\label{alg: align-first-cycle}
\caption{First cycle of the alignment procedure.}
\textbf{Input}: Sentence $s$; Scene graph $G(s)$ \\
Initialize aligned nodes $N$ as empty set \\
Initialize aligned words $W$ as empty set \\
\For{$o$ in object nodes of $G(s) \setminus N$}{
\For{$w$ in $s \setminus W$}{
\If{$o \leftrightarrow w$ according to WBW}{
Add $(o, w)$; $N = N \cup \{o\}$; $W = W \cup \{w\}$
}
}
}
\For{$a$ in attribute nodes of $G(s) \setminus N$}{
\For{$w$ in $s \setminus W$}{
\If{$a \leftrightarrow w$ according to WBW or SYN \textbf{and} $a$'s object node is in $N$}{
Add $(a, w)$; $N = N \cup \{a\}$; $W = W \cup \{w \}$
}
}
}
\For{$r$ in relation nodes of $G(s) \setminus N$}{
\For{$w$ in $s \setminus W$}{
\If{$r \leftrightarrow w$ according to WBW or SYN \textbf{and} $r$'s subject and object nodes are both in $N$}{
Add $(r, w)$; $N = N \cup \{r \}$; $W = W \cup \{w\}$
}
}
}
\end{algorithm}

%% file: parsing.tex
\section{Customized Dependency Parsing}

In the previous section, we have established the connection between scene graph generation and dependency parsing, which assigns a parent word for every word in a sentence, as well as a label for this directed arc.
We start by describing our base dependency parsing model, which is neural network based and performs among the state-of-the-art.
We then show why and how we do customization, such that scene graph generation is achieved with a single, end-to-end model.

\subsection{Neural Dependency Parsing Base Model}

We base our model on the transition-based parser of \citet{DBLP:journals/tacl/KiperwasserG16}.
Here we describe its key components: the arc-hybrid system that defines the transition actions, the neural architecture for feature extractor and scoring function, and the loss function.

\paragraph{The Arc-Hybrid System}
In the arc-hybrid system, a configuration consists of a stack $\sigma$, a buffer $\beta$, and a set $T$ of dependency arcs.
Given a sentence $s = w_1, \hdots, w_n$, the system is initialized with an empty stack $\sigma$, an empty arc set $T$, and $\beta = 1, \hdots, n, \texttt{ROOT}$, where \texttt{ROOT} is a special index.
The system terminates when $\sigma$ is empty and $\beta$ contains only \texttt{ROOT}.
The dependency tree is given by the arc set $T$ upon termination.

\begin{table*}[t]
\centering
\begin{tabular}{ccc | c | ccc}
\toprule
\bf{Stack} $\sigma_t$ & \bf{Buffer} $\beta_t$ & \bf{Arc set} $T_t$ & \bf{Action} & \bf{Stack} $\sigma_{t+1}$ & \bf{Buffer} $\beta_{t+1}$ & \bf{Arc set} $T_{t+1}$ \\
\midrule
$\sigma$ & $b_0 | \beta$ & $T$ & \textsc{Shift} & $\sigma | b_0$ & $\beta$ & $T$ \\
$\sigma | s_1 | s_0$ & $b_0 | \beta$ & $T$ & \textsc{Left($l$)} & $\sigma | s_1$ & $b_0 | \beta$ & $T \cup \{ (b_0, s_0, l) \}$ \\
$\sigma | s_1 | s_0$ & $\beta$ & $T$ & \textsc{Right($l$)} & $\sigma | s_1$ & $\beta$ & $T \cup \{ (s_1, s_0, l) \}$\\
\midrule
$\sigma | s_0$ & $\beta$ & $T$ & \textsc{Reduce} & $\sigma$ & $\beta$ & $T$ \\
\bottomrule
\end{tabular}
\caption{Transition actions under the arc-hybrid system. The first three actions are from dependency parsing; the last one is introduced for scene graph parsing.}
\label{tab: actions}
\end{table*}

The arc-hybrid system allows three transition actions, \textsc{Shift}, \textsc{Left$_l$}, \textsc{Right$_l$}, described in Table~\ref{tab: actions}.
The \textsc{Shift} transition moves the first element of the buffer to the stack.
The \textsc{Left($l$)} transition yields an arc from the first element of the buffer to the top element of the stack, and then removes the top element from the stack.
The \textsc{Right($l$)} transition yields an arc from the second top element of the stack to the top element of the stack, and then also removes the top element from the stack.

The following paragraphs describe how to select the correct transition action (and label $l$) in each step in order to generate a correct dependency tree.

\paragraph{BiLSTM Feature Extractor}

Let the word embeddings of a sentence $s$ be $\mathbf{w}_1, \hdots, \mathbf{w}_n$.
An LSTM cell is a parameterized function that takes as input $\mathbf{w}_t$, and updates its hidden states:
\begin{equation}
\text{LSTM cell}: (\mathbf{w}_t, \mathbf{h}_{t-1}) \rightarrow \mathbf{h}_t
\end{equation}
As a result, an LSTM network, which simply applies the LSTM cell $t$ times, is a parameterized function mapping a sequence of input vectors $\mathbf{w}_{1:t}$ to a sequence of output vectors $\mathbf{h}_{1:t}$.
In our notation, we drop the intermediate vectors $\mathbf{h}_{1:t-1}$ and let $\textsc{LSTM}(\mathbf{w}_{1:t})$ represent $\mathbf{h}_t$.

A bidirectional LSTM, or BiLSTM for short, consists of two LSTMs: $\textsc{LSTM}_F$ which reads the input sequence in the original order, and $\textsc{LSTM}_B$ which reads it in reverse.
Then
\begin{align}
\textsc{BiLSTM}&(\mathbf{w}_{1:n}, i) = \nonumber \\
&\textsc{LSTM}_F(\mathbf{w}_{1:i}) \circ \textsc{LSTM}_B(\mathbf{w}_{n:i})
\end{align}
where $\circ$ denotes concatenation.
Intuitively, the forward LSTM encodes information from the left side of the $i$-th word and the backward LSTM encodes information to its right, such that the vector $\mathbf{v}_i = \textsc{BiLSTM}(\mathbf{w}_{1:n}, i)$ has the full sentence as context.

When predicting the transition action, the feature function $\phi(c)$ that summarizes the current configuration $c = (\sigma, \beta, T)$ is simply the concatenated BiLSTM vectors of the top three elements in the stack and the first element in the buffer:
\begin{equation}
\phi(c) = \mathbf{v}_{s_2} \circ \mathbf{v}_{s_1} \circ \mathbf{v}_{s_0} \circ \mathbf{v}_{b_0}
\end{equation}

\paragraph{MLP Scoring Function}

The score of transition action $y$ under the current configuration $c$ is determined by a multi-layer perceptron with one hidden layer:
\begin{equation}
f(c, y) = MLP(\phi(c))[y] \label{eqn: score}
\end{equation}
where
\begin{equation}
MLP(x) = W_2 \cdot \tanh(W_1 \cdot x + b_1) + b_2
\end{equation}

\paragraph{Hinge Loss Function}
The training objective is to raise the scores of correct transitions above scores of incorrect ones.
Therefore, at each step, we use a hinge loss defined as:
\begin{align}
\mathcal{L} = \max(0, 1 &- \max_{y^+ \in Y^+} f(c, y^+) \nonumber \\
&+ \max_{y^- \in Y \setminus Y^+} f(c, y^-)) \label{eqn: loss}
\end{align}
where $Y$ is the set of possible transitions and $Y^+$ is the set of correct transitions at the current step.
In each training step, the parser scores all possible transitions using Eqn.~\ref{eqn: score}, incurs a loss using Eqn.~\ref{eqn: loss}, selects a following transition, and updates the configuration.
Losses at individual steps are summed throughout the parsing of a sentence, and then parameters are updated using backpropagation.

In test time, we simply choose the transition action that yields the highest score at each step.

\subsection{Customization}

In order to generate scene graphs with dependency parsing, modification is necessary for at least two reasons.
First, we need to redefine the label space of arcs so as to reflect the edge-centric representation of a scene graph.
Second, not every word in the sentence will be (part of) a node in the scene graph (see Figure~\ref{fig: qualitative-example} for an example). 
In other words, some words in the sentence may not have a parent word, which violates the dependency parsing setting.
We tackle these two challenges by redesigning the edge labels and expanding the set of transition actions.

\paragraph{Redesigning Edge Labels}

We define a total of five edge labels, so as to faithfully bridge the edge-centric view of scene graphs with dependency parsing models:
\begin{itemize}
\item \texttt{CONT}: This label is created for nodes whose label is a phrase.
For example, the phrase "in front of" is a single relation node in the scene graph.
By introducing the \texttt{CONT} label, we expect the parsing result to be either 
\begin{equation}
\text{in} \xrightarrow{\texttt{CONT}} \text{front} \xrightarrow{\texttt{CONT}} \text{of}
\end{equation}
or
\begin{equation}
\text{in} \xleftarrow{\texttt{CONT}} \text{front} \xleftarrow{\texttt{CONT}} \text{of}
\end{equation}
where the direction of the arcs (left or right) is predefined by hand.

The leftmost word under the right arc rule or the rightmost word under the left arc rule is called the \textit{head} of the phrase.
A single-word node does not need this \texttt{CONT} label, and the head is itself.
\item \texttt{ATTR}: The arc label from the head of an object node to the head of an attribute node.
\item \texttt{SUBJ}: The arc label from the head of an object node (subject) to the head of a relation node.
\item \texttt{OBJT}: The arc label from the head of a relation node to the head of an object node (object).
\item \texttt{BEGN}: The arc label from the \texttt{ROOT} index to all heads of object nodes without a parent.
\end{itemize}

\paragraph{Expanding Transition Actions}

With the three transition actions \textsc{Shift}, \textsc{Left($l$)}, \textsc{Right($l$)}, we only drop an element (from the top of the stack) after it has already been associated with an arc.
This design ensures that an arc is associated with every word.
However, in our setting for scene graph generation, there may be no arc for some of the words, especially empty words.

Our solution is to augment the action set with a \textsc{Reduce} action, that pops the stack \textit{without} adding to the arc set (see Table~\ref{tab: actions}).
This action is often used in other transition-based dependency parsing systems (e.g. arc-eager \citep{nivre2004incrementality}).
More recently, \citet{DBLP:conf/acl/HershcovichAR17} and \citet{DBLP:conf/acl/BuysB17} also included this action when parsing sentences to graph structures.

We still minimize the loss function defined in Eqn.~\ref{eqn: loss}, except that now $|Y|$ increases from 3 to 4.
During training, we impose the oracle to select the \textsc{Reduce} action when it is in $Y^+$.
In terms of loss function, we increment by 1 the loss incurred by the other 3 transition actions if \textsc{Reduce} incurs zero loss. 



%% file: exp.tex
\section{Experiments}

\subsection{Implementation Details}

We train and evaluate our scene graph parsing model on (a subset of) the Visual Genome \citep{DBLP:journals/ijcv/KrishnaZGJHKCKL17} dataset. 
Each image in Visual Genome contains a number of regions, and each region is annotated with both a region description and a region scene graph.
Our training set is the intersection of Visual Genome and MS COCO \citep{DBLP:conf/eccv/LinMBHPRDZ14} train2014 set, which contains a total of 34027 images/ 1070145 regions.
We evaluate on the intersection of Visual Genome and MS COCO val2014 set, which contains a total of 17471 images/ 547795 regions.

In our experiments, the number of hidden units in BiLSTM is 256; the number of layers in BiLSTM is 2; the word embedding dimension is 200; the number of hidden units in MLP is 100.
We use fixed learning rate 0.001 and Adam optimizer \citep{DBLP:journals/corr/KingmaB14} with epsilon 0.01.
Training usually converges within 4 epochs.

We will release our code and trained model upon acceptance.

\begin{table}[t]
\centering
\begin{tabular}{ll}
\toprule
{\bf Parser} & {\bf F-score} \\
\midrule
Stanford \cite{schuster2015generating} & 0.3549 \\
SPICE \cite{DBLP:conf/eccv/AndersonFJG16} & 0.4469 \\
\midrule
Ours (left arc rule) & {\bf 0.4967} \\
Ours (right arc rule) & 0.4952 \\
Ours (all SYN) & 0.4877 \\
Ours (no SYN) & 0.4538 \\
\midrule
Oracle & 0.6985 \\
\bottomrule
\end{tabular}
\caption{The F-scores (i.e. SPICE metric) between scene graphs parsed from region descriptions and ground truth region graphs on the intersection of Visual Genome \cite{DBLP:journals/ijcv/KrishnaZGJHKCKL17} and MS COCO \cite{DBLP:conf/eccv/LinMBHPRDZ14} validation set.}
\label{tab: fscore}
\end{table}

\begin{figure*}[t]
\centering
\includegraphics[width=\textwidth]{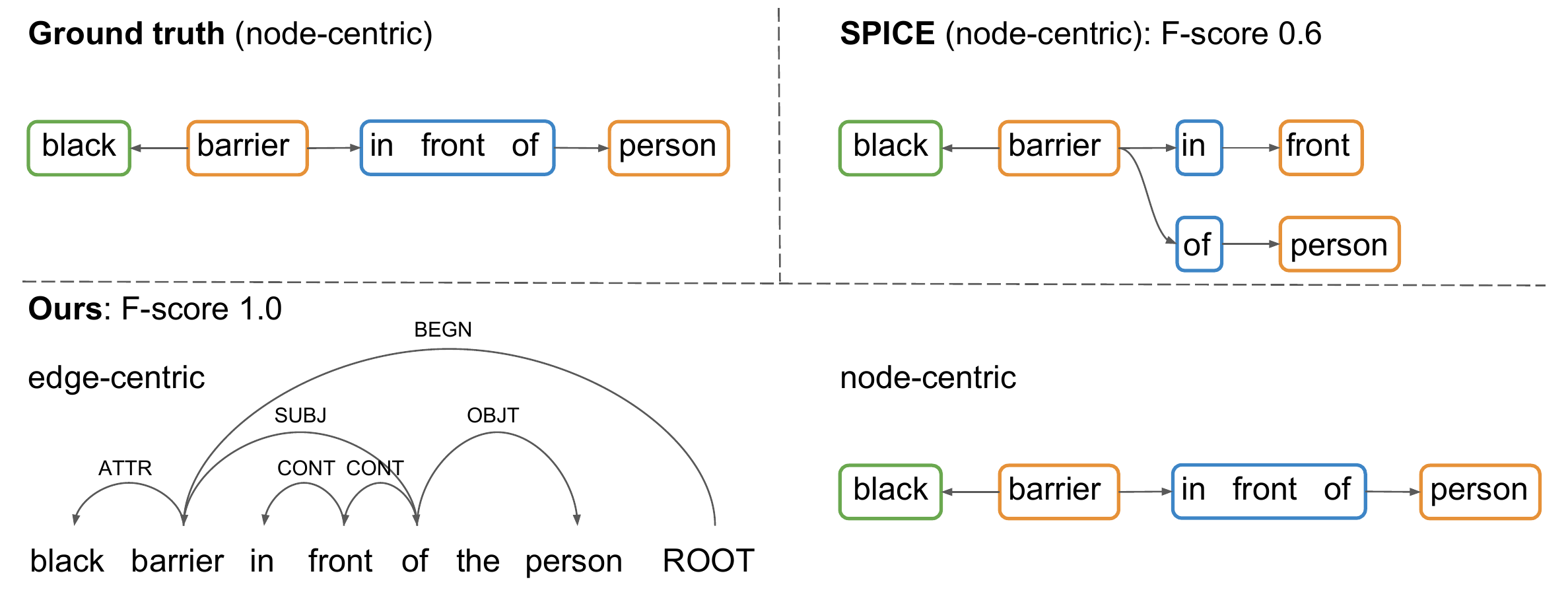}
\caption{Scene graph parsing result of the sentence ``black barrier in front of the person''.
In the node-centric graphs, orange represents object node, green represents attribute node, blue represents relation node.}
\label{fig: qualitative-example}
\end{figure*}

\begin{figure*}[t]
\centering
\begin{tabular}{l | l | l | l}
\toprule
 & {\bf Stack} & {\bf Buffer} & {\bf Action} \\
\midrule
0 & & black barrier in front of the person \texttt{ROOT} & \textsc{Shift} \\
1 & black & barrier in front of the person \texttt{ROOT} & \textsc{Left}(\texttt{ATTR}) \\
2 & & barrier in front of the person \texttt{ROOT} & \textsc{Shift} \\
3 & barrier & in front of the person \texttt{ROOT} & \textsc{Shift} \\
4 & barrier in & front of the person \texttt{ROOT} & \textsc{Left}(\texttt{CONT}) \\
5 & barrier & front of the person \texttt{ROOT} & \textsc{Shift} \\
6 & barrier front & of the person \texttt{ROOT} & \textsc{Left}(\texttt{CONT}) \\
7 & barrier & of the person \texttt{ROOT} & \textsc{Shift} \\
8 & barrier of & the person \texttt{ROOT} & \textsc{Shift} \\
9 & barrier of the & person \texttt{ROOT} & \textsc{Reduce} \\
10 & barrier of & person \texttt{ROOT} & \textsc{Shift} \\
11 & barrier of person & \texttt{ROOT} & \textsc{Right}(\texttt{OBJT}) \\
12 & barrier of & \texttt{ROOT} & \textsc{Right}(\texttt{SUBJ}) \\
13 & barrier & \texttt{ROOT} & \textsc{Left}(\texttt{BEGN}) \\
14 & & \texttt{ROOT} & \\
\bottomrule
\end{tabular}
\caption{Intermediate actions taken by the trained dependency parser when parsing the sentence ``black barrier in front of the person''.}
\label{fig: parse-example}
\end{figure*}

\begin{table*}[t]
\centering
\begin{tabular}{l|ccc|ccc}
\toprule
& \multicolumn{3}{c|}{{\bf Development set}} & \multicolumn{3}{c}{{\bf Test set}} \\
\midrule
& {\bf R@5} & {\bf R@10} & {\bf Med. rank} & {\bf R@5} & {\bf R@10} & {\bf Med. rank} \\
\midrule
\cite{schuster2015generating} & 33.82\% & 45.58\% & 6 & 34.96\% & 45.68\% & {\bf 5} \\
Ours & {\bf 36.69\%} & {\bf 49.41\%} & {\bf 4} & {\bf 36.70\%} & {\bf 49.37\%} & {\bf 5} \\
\bottomrule
\end{tabular}
\caption{Image retrieval results. 
We follow the same experiment setup as \citet{schuster2015generating}, except using a different scoring function when ranking images.
Our parser consistently outperforms the Stanford Scene Graph Parser across evaluation metrics.}
\label{tab: retrieval}
\end{table*}

\subsection{Quality of Parsed Scene Graphs}

We use a slightly modified version of SPICE score \citep{DBLP:conf/eccv/AndersonFJG16} to evaluate the quality of scene graph parsing.
Specifically, for every region, we parse its description using a parser (e.g. the one used in SPICE or our customized dependency parser), and then calculate the F-score between the parsed graph and the ground truth region graph (see Section 3.2 of \citet{DBLP:conf/eccv/AndersonFJG16} for more details).
Note that when SPICE calculates the F-score, a node in one graph could be matched to several nodes in the other, which is problematic.
We fix this and enforce one-to-one matching when calculating the F-score.
Finally, we report the average F-score across all regions.

Table~\ref{tab: fscore} summarizes our results.
We see that our customized dependency parsing model achieves an average F-score of 49.67\%, which significantly outperforms the parser used in SPICE by 5 percent.
This result shows that our customized dependency parser is very effective at learning from data, and generates more accurate scene graphs than the best previous approach. 

\paragraph{Ablation Studies}
First, we study how the sentence-graph alignment procedure affects the final performance.
Recall that our procedure involves two cycles, each with three steps.
Of the six steps, synonym match (SYN) is only not used in the first step.
We tried two more settings, where SYN is either used in all six steps or none of the six steps.
We can see from Table~\ref{tab: fscore} that the final F-score drops in both cases, hence supporting the procedure that we chose.

Second, we study whether changing the direction of \texttt{CONT} arcs from pointing left to pointing right will make much difference.
Table~\ref{tab: fscore} shows that the two choices give very similar performance, suggesting that our dependency parser is robust to this design choice.

Finally, we report the oracle score, which is the similarity between the aligned graphs that we use during training and the ground truth graphs.
The F-score is relatively high at 69.85\%.
This shows that improving the parser (about 20\% margin) and improving the sentence-graph alignment (about 30\% margin) are both promising directions for future research.

\paragraph{Qualitative Examples}
We provide one parsing example in Figure~\ref{fig: qualitative-example} and Figure~\ref{fig: parse-example}.
This is a sentence that is relatively simple, and the underlying scene graph includes two object nodes, one attribute node, and one compound word relation node.
In parsing this sentence, all four actions listed in Table~\ref{tab: actions} are used (see Figure~\ref{fig: parse-example}) to produce the edge-centric scene graph (bottom left of Figure~\ref{fig: qualitative-example}), which is then trivially converted to the node-centric scene graph (bottom right of Figure~\ref{fig: qualitative-example}). 

\subsection{Application in Image Retrieval}

We test if the advantage of our parser can be propagated to computer vision tasks, such as image retrieval.
We directly compare our parser with the Stanford Scene Graph Parser \citep{schuster2015generating} on the development set and test set of the image retrieval dataset used in \citet{schuster2015generating} (not Visual Genome).

For every region in an image, there is a human-annotated region description and region scene graph. 
The queries are the region descriptions. 
If the region graph corresponding to the query is a subgraph of the complete graph of another image, then that image is added to the ground truth set for this query. 
All these are strictly following \citet{schuster2015generating}.
However, since we did not obtain nor reproduce the CRF model used in \citet{DBLP:conf/cvpr/JohnsonKSLSBL15} and \citet{schuster2015generating}, we used F-score similarity instead of the likelihood of the maximum a posteriori CRF solution when ranking the images based on the region descriptions.
Therefore the numbers we report in Table~\ref{tab: retrieval} are not directly comparable with those reported in \citet{schuster2015generating}.

Our parser delivers better retrieval performance across all three evaluation metrics: recall@5, recall@10, and median rank.
We also notice that the numbers in our retrieval setting are higher than those (even with oracle) in \citet{schuster2015generating}'s retrieval setting.
This strongly suggests that generating accurate scene graphs from images is a very promising research direction in image retrieval, and grounding parsed scene graphs to bounding box proposals without considering visual attributes/relationships \citep{DBLP:conf/cvpr/JohnsonKSLSBL15} is suboptimal.


%% file: conc.tex
\section{Conclusion}

In this paper, we offer a new perspective and solution to the task of parsing scene graphs from textual descriptions.
We begin by moving the labels/types from the nodes to the edges and introducing the edge-centric view of scene graphs.
We further show that the gap between edge-centric scene graphs and dependency parses can be filled with a careful redesign of label and action space.
This motivates us to train a single, customized, end-to-end neural dependency parser for this task, as opposed to prior approaches that used generic dependency parsing followed by heuristics or simple classifier.
We directly train our parser on a subset of Visual Genome \citep{DBLP:journals/ijcv/KrishnaZGJHKCKL17}, without transferring any knowledge from Penn Treebank \citep{DBLP:journals/coling/MarcusSM94} as previous works 
did.
The quality of our trained parser is validated in terms of both SPICE similarity to the ground truth graphs and recall rate/median rank when performing image retrieval.

We hope our paper can lead to more thoughts on the creative uses and extensions of existing NLP tools to tasks and datasets in other domains.
In the future, we plan to tackle more computer vision tasks with this improved scene graph parsing technique in hand, such as image region grounding.
We also plan to investigate parsing scene graph with cyclic structures, as well as whether/how the image information can help boost parsing quality.

%% file: sgparser.bbl
\begin{thebibliography}{}
\expandafter\ifx\csname natexlab\endcsname\relax\def\natexlab#1{#1}\fi

\bibitem[{Anderson et~al.(2016)Anderson, Fernando, Johnson, and
  Gould}]{DBLP:conf/eccv/AndersonFJG16}
Peter Anderson, Basura Fernando, Mark Johnson, and Stephen Gould. 2016.
\newblock {SPICE:} semantic propositional image caption evaluation.
\newblock In {\em {ECCV}\/}. Springer, volume 9909 of {\em Lecture Notes in
  Computer Science\/}, pages 382--398.

\bibitem[{Andreas et~al.(2016)Andreas, Rohrbach, Darrell, and
  Klein}]{DBLP:conf/cvpr/AndreasRDK16}
Jacob Andreas, Marcus Rohrbach, Trevor Darrell, and Dan Klein. 2016.
\newblock Neural module networks.
\newblock In {\em {CVPR}\/}. {IEEE} Computer Society, pages 39--48.

\bibitem[{Antol et~al.(2015)Antol, Agrawal, Lu, Mitchell, Batra, Zitnick, and
  Parikh}]{DBLP:conf/iccv/AntolALMBZP15}
Stanislaw Antol, Aishwarya Agrawal, Jiasen Lu, Margaret Mitchell, Dhruv Batra,
  C.~Lawrence Zitnick, and Devi Parikh. 2015.
\newblock {VQA:} visual question answering.
\newblock In {\em {ICCV}\/}. {IEEE} Computer Society, pages 2425--2433.

\bibitem[{Buys and Blunsom(2017)}]{DBLP:conf/acl/BuysB17}
Jan Buys and Phil Blunsom. 2017.
\newblock Robust incremental neural semantic graph parsing.
\newblock In {\em {ACL}\/}. Association for Computational Linguistics, pages
  1215--1226.

\bibitem[{Chang et~al.(2014)Chang, Savva, and
  Manning}]{DBLP:conf/emnlp/ChangSM14}
Angel~X. Chang, Manolis Savva, and Christopher~D. Manning. 2014.
\newblock Learning spatial knowledge for text to 3d scene generation.
\newblock In {\em {EMNLP}\/}. {ACL}, pages 2028--2038.

\bibitem[{Chen and Manning(2014)}]{DBLP:conf/emnlp/ChenM14}
Danqi Chen and Christopher~D. Manning. 2014.
\newblock A fast and accurate dependency parser using neural networks.
\newblock In {\em {EMNLP}\/}. {ACL}, pages 740--750.

\bibitem[{Cho et~al.(2014)Cho, van Merrienboer, G{\"{u}}l{\c{c}}ehre, Bahdanau,
  Bougares, Schwenk, and Bengio}]{DBLP:conf/emnlp/ChoMGBBSB14}
Kyunghyun Cho, Bart van Merrienboer, {\c{C}}aglar G{\"{u}}l{\c{c}}ehre, Dzmitry
  Bahdanau, Fethi Bougares, Holger Schwenk, and Yoshua Bengio. 2014.
\newblock Learning phrase representations using {RNN} encoder-decoder for
  statistical machine translation.
\newblock In {\em {EMNLP}\/}. {ACL}, pages 1724--1734.

\bibitem[{Cross and Huang(2016)}]{DBLP:conf/acl/CrossH16}
James Cross and Liang Huang. 2016.
\newblock Incremental parsing with minimal features using bi-directional
  {LSTM}.
\newblock In {\em {ACL} {(2)}\/}. The Association for Computer Linguistics.

\bibitem[{Dai et~al.(2017)Dai, Zhang, and Lin}]{DBLP:conf/cvpr/DaiZL17}
Bo~Dai, Yuqi Zhang, and Dahua Lin. 2017.
\newblock Detecting visual relationships with deep relational networks.
\newblock In {\em {CVPR}\/}. {IEEE} Computer Society, pages 3298--3308.

\bibitem[{Denkowski and Lavie(2014)}]{DBLP:conf/wmt/DenkowskiL14}
Michael~J. Denkowski and Alon Lavie. 2014.
\newblock Meteor universal: Language specific translation evaluation for any
  target language.
\newblock In {\em WMT@ACL\/}. The Association for Computer Linguistics, pages
  376--380.

\bibitem[{Donahue et~al.(2015)Donahue, Hendricks, Guadarrama, Rohrbach,
  Venugopalan, Darrell, and Saenko}]{DBLP:conf/cvpr/DonahueHGRVDS15}
Jeff Donahue, Lisa~Anne Hendricks, Sergio Guadarrama, Marcus Rohrbach,
  Subhashini Venugopalan, Trevor Darrell, and Kate Saenko. 2015.
\newblock Long-term recurrent convolutional networks for visual recognition and
  description.
\newblock In {\em {CVPR}\/}. {IEEE} Computer Society, pages 2625--2634.

\bibitem[{Donahue et~al.(2014)Donahue, Jia, Vinyals, Hoffman, Zhang, Tzeng, and
  Darrell}]{DBLP:conf/icml/DonahueJVHZTD14}
Jeff Donahue, Yangqing Jia, Oriol Vinyals, Judy Hoffman, Ning Zhang, Eric
  Tzeng, and Trevor Darrell. 2014.
\newblock Decaf: {A} deep convolutional activation feature for generic visual
  recognition.
\newblock In {\em {ICML}\/}. JMLR.org, volume~32 of {\em {JMLR} Workshop and
  Conference Proceedings\/}, pages 647--655.

\bibitem[{Dozat and Manning(2016)}]{DBLP:journals/corr/DozatM16}
Timothy Dozat and Christopher~D. Manning. 2016.
\newblock Deep biaffine attention for neural dependency parsing.
\newblock {\em CoRR\/} abs/1611.01734.

\bibitem[{Dyer et~al.(2015)Dyer, Ballesteros, Ling, Matthews, and
  Smith}]{DBLP:conf/acl/DyerBLMS15}
Chris Dyer, Miguel Ballesteros, Wang Ling, Austin Matthews, and Noah~A. Smith.
  2015.
\newblock Transition-based dependency parsing with stack long short-term
  memory.
\newblock In {\em {ACL} {(1)}\/}. The Association for Computer Linguistics,
  pages 334--343.

\bibitem[{Flanigan et~al.(2014)Flanigan, Thomson, Carbonell, Dyer, and
  Smith}]{DBLP:conf/acl/FlaniganTCDS14}
Jeffrey Flanigan, Sam Thomson, Jaime~G. Carbonell, Chris Dyer, and Noah~A.
  Smith. 2014.
\newblock A discriminative graph-based parser for the abstract meaning
  representation.
\newblock In {\em {ACL} {(1)}\/}. The Association for Computer Linguistics,
  pages 1426--1436.

\bibitem[{Frome et~al.(2013)Frome, Corrado, Shlens, Bengio, Dean, Ranzato, and
  Mikolov}]{DBLP:conf/nips/FromeCSBDRM13}
Andrea Frome, Gregory~S. Corrado, Jonathon Shlens, Samy Bengio, Jeffrey Dean,
  Marc'Aurelio Ranzato, and Tomas Mikolov. 2013.
\newblock Devise: {A} deep visual-semantic embedding model.
\newblock In {\em {NIPS}\/}. pages 2121--2129.

\bibitem[{Hershcovich et~al.(2017)Hershcovich, Abend, and
  Rappoport}]{DBLP:conf/acl/HershcovichAR17}
Daniel Hershcovich, Omri Abend, and Ari Rappoport. 2017.
\newblock A transition-based directed acyclic graph parser for {UCCA}.
\newblock In {\em {ACL}\/}. Association for Computational Linguistics, pages
  1127--1138.

\bibitem[{Hochreiter and Schmidhuber(1997)}]{DBLP:journals/neco/HochreiterS97}
Sepp Hochreiter and J{\"{u}}rgen Schmidhuber. 1997.
\newblock Long short-term memory.
\newblock {\em Neural Computation\/} 9(8):1735--1780.

\bibitem[{Hu et~al.(2016)Hu, Xu, Rohrbach, Feng, Saenko, and
  Darrell}]{DBLP:conf/cvpr/HuXRFSD16}
Ronghang Hu, Huazhe Xu, Marcus Rohrbach, Jiashi Feng, Kate Saenko, and Trevor
  Darrell. 2016.
\newblock Natural language object retrieval.
\newblock In {\em {CVPR}\/}. {IEEE} Computer Society, pages 4555--4564.

\bibitem[{Johnson et~al.(2015)Johnson, Krishna, Stark, Li, Shamma, Bernstein,
  and Li}]{DBLP:conf/cvpr/JohnsonKSLSBL15}
Justin Johnson, Ranjay Krishna, Michael Stark, Li{-}Jia Li, David~A. Shamma,
  Michael~S. Bernstein, and Fei{-}Fei Li. 2015.
\newblock Image retrieval using scene graphs.
\newblock In {\em {CVPR}\/}. {IEEE} Computer Society, pages 3668--3678.

\bibitem[{Kaiser et~al.(2017)Kaiser, Gomez, Shazeer, Vaswani, Parmar, Jones,
  and Uszkoreit}]{DBLP:journals/corr/KaiserGSVPJU17}
Lukasz Kaiser, Aidan~N. Gomez, Noam Shazeer, Ashish Vaswani, Niki Parmar, Llion
  Jones, and Jakob Uszkoreit. 2017.
\newblock One model to learn them all.
\newblock {\em CoRR\/} abs/1706.05137.

\bibitem[{Karpathy and Li(2015)}]{DBLP:conf/cvpr/KarpathyL15}
Andrej Karpathy and Fei{-}Fei Li. 2015.
\newblock Deep visual-semantic alignments for generating image descriptions.
\newblock In {\em {CVPR}\/}. {IEEE} Computer Society, pages 3128--3137.

\bibitem[{Katiyar and Cardie(2017)}]{DBLP:conf/acl/KatiyarC17}
Arzoo Katiyar and Claire Cardie. 2017.
\newblock Going out on a limb: Joint extraction of entity mentions and
  relations without dependency trees.
\newblock In {\em {ACL}\/}. Association for Computational Linguistics, pages
  917--928.

\bibitem[{Kingma and Ba(2014)}]{DBLP:journals/corr/KingmaB14}
Diederik~P. Kingma and Jimmy Ba. 2014.
\newblock Adam: {A} method for stochastic optimization.
\newblock {\em CoRR\/} abs/1412.6980.

\bibitem[{Kiperwasser and Goldberg(2016)}]{DBLP:journals/tacl/KiperwasserG16}
Eliyahu Kiperwasser and Yoav Goldberg. 2016.
\newblock Simple and accurate dependency parsing using bidirectional {LSTM}
  feature representations.
\newblock {\em {TACL}\/} 4:313--327.

\bibitem[{Konstas et~al.(2017)Konstas, Iyer, Yatskar, Choi, and
  Zettlemoyer}]{DBLP:conf/acl/KonstasIYCZ17}
Ioannis Konstas, Srinivasan Iyer, Mark Yatskar, Yejin Choi, and Luke
  Zettlemoyer. 2017.
\newblock Neural {AMR:} sequence-to-sequence models for parsing and generation.
\newblock In {\em {ACL} {(1)}\/}. Association for Computational Linguistics,
  pages 146--157.

\bibitem[{Krishna et~al.(2017)Krishna, Zhu, Groth, Johnson, Hata, Kravitz,
  Chen, Kalantidis, Li, Shamma, Bernstein, and
  Fei{-}Fei}]{DBLP:journals/ijcv/KrishnaZGJHKCKL17}
Ranjay Krishna, Yuke Zhu, Oliver Groth, Justin Johnson, Kenji Hata, Joshua
  Kravitz, Stephanie Chen, Yannis Kalantidis, Li{-}Jia Li, David~A. Shamma,
  Michael~S. Bernstein, and Li~Fei{-}Fei. 2017.
\newblock Visual genome: Connecting language and vision using crowdsourced
  dense image annotations.
\newblock {\em International Journal of Computer Vision\/} 123(1):32--73.

\bibitem[{Krishnamurthy and Kollar(2013)}]{DBLP:journals/tacl/KrishnamurthyK13}
Jayant Krishnamurthy and Thomas Kollar. 2013.
\newblock Jointly learning to parse and perceive: Connecting natural language
  to the physical world.
\newblock {\em {TACL}\/} 1:193--206.

\bibitem[{K{\"{u}}bler et~al.(2009)K{\"{u}}bler, McDonald, and
  Nivre}]{DBLP:series/synthesis/2009Kubler}
Sandra K{\"{u}}bler, Ryan~T. McDonald, and Joakim Nivre. 2009.
\newblock {\em Dependency Parsing\/}.
\newblock Synthesis Lectures on Human Language Technologies. Morgan {\&}
  Claypool Publishers.

\bibitem[{Li et~al.(2017)Li, Ouyang, Zhou, Wang, and
  Wang}]{DBLP:journals/corr/LiOZWW17}
Yikang Li, Wanli Ouyang, Bolei Zhou, Kun Wang, and Xiaogang Wang. 2017.
\newblock Scene graph generation from objects, phrases and caption regions.
\newblock {\em CoRR\/} abs/1707.09700.

\bibitem[{Lin et~al.(2014)Lin, Maire, Belongie, Hays, Perona, Ramanan,
  Doll{\'{a}}r, and Zitnick}]{DBLP:conf/eccv/LinMBHPRDZ14}
Tsung{-}Yi Lin, Michael Maire, Serge~J. Belongie, James Hays, Pietro Perona,
  Deva Ramanan, Piotr Doll{\'{a}}r, and C.~Lawrence Zitnick. 2014.
\newblock Microsoft {COCO:} common objects in context.
\newblock In {\em {ECCV}\/}. Springer, volume 8693 of {\em Lecture Notes in
  Computer Science\/}, pages 740--755.

\bibitem[{Liu et~al.(2017{\natexlab{a}})Liu, Lin, Shen, Yang, Lu, and
  Yuille}]{DBLP:conf/iccv/LiuLSYLY17}
Chenxi Liu, Zhe Lin, Xiaohui Shen, Jimei Yang, Xin Lu, and Alan~L. Yuille.
  2017{\natexlab{a}}.
\newblock Recurrent multimodal interaction for referring image segmentation.
\newblock In {\em {ICCV}\/}. {IEEE} Computer Society, pages 1280--1289.

\bibitem[{Liu et~al.(2017{\natexlab{b}})Liu, Mao, Sha, and
  Yuille}]{DBLP:conf/aaai/LiuMSY17}
Chenxi Liu, Junhua Mao, Fei Sha, and Alan~L. Yuille. 2017{\natexlab{b}}.
\newblock Attention correctness in neural image captioning.
\newblock In {\em {AAAI}\/}. {AAAI} Press, pages 4176--4182.

\bibitem[{Lu et~al.(2016)Lu, Krishna, Bernstein, and
  Li}]{DBLP:conf/eccv/LuKBL16}
Cewu Lu, Ranjay Krishna, Michael~S. Bernstein, and Fei{-}Fei Li. 2016.
\newblock Visual relationship detection with language priors.
\newblock In {\em {ECCV}\/}. Springer, volume 9905 of {\em Lecture Notes in
  Computer Science\/}, pages 852--869.

\bibitem[{Mao et~al.(2016)Mao, Huang, Toshev, Camburu, Yuille, and
  Murphy}]{DBLP:conf/cvpr/MaoHTCY016}
Junhua Mao, Jonathan Huang, Alexander Toshev, Oana Camburu, Alan~L. Yuille, and
  Kevin Murphy. 2016.
\newblock Generation and comprehension of unambiguous object descriptions.
\newblock In {\em {CVPR}\/}. {IEEE} Computer Society, pages 11--20.

\bibitem[{Mao et~al.(2014)Mao, Xu, Yang, Wang, and
  Yuille}]{DBLP:journals/corr/MaoXYWY14a}
Junhua Mao, Wei Xu, Yi~Yang, Jiang Wang, and Alan~L. Yuille. 2014.
\newblock Deep captioning with multimodal recurrent neural networks (m-rnn).
\newblock {\em CoRR\/} abs/1412.6632.

\bibitem[{Marcus et~al.(1993)Marcus, Santorini, and
  Marcinkiewicz}]{DBLP:journals/coling/MarcusSM94}
Mitchell~P. Marcus, Beatrice Santorini, and Mary~Ann Marcinkiewicz. 1993.
\newblock Building a large annotated corpus of english: The penn treebank.
\newblock {\em Computational Linguistics\/} 19(2):313--330.

\bibitem[{Mikolov et~al.(2013)Mikolov, Chen, Corrado, and
  Dean}]{DBLP:journals/corr/abs-1301-3781}
Tomas Mikolov, Kai Chen, Greg Corrado, and Jeffrey Dean. 2013.
\newblock Efficient estimation of word representations in vector space.
\newblock {\em CoRR\/} abs/1301.3781.

\bibitem[{Nivre(2004)}]{nivre2004incrementality}
Joakim Nivre. 2004.
\newblock Incrementality in deterministic dependency parsing.
\newblock In {\em Proceedings of the Workshop on Incremental Parsing: Bringing
  Engineering and Cognition Together\/}. Association for Computational
  Linguistics, pages 50--57.

\bibitem[{Pennington et~al.(2014)Pennington, Socher, and
  Manning}]{DBLP:conf/emnlp/PenningtonSM14}
Jeffrey Pennington, Richard Socher, and Christopher~D. Manning. 2014.
\newblock Glove: Global vectors for word representation.
\newblock In {\em {EMNLP}\/}. {ACL}, pages 1532--1543.

\bibitem[{Schuster et~al.(2015)Schuster, Krishna, Chang, Fei-Fei, and
  Manning}]{schuster2015generating}
Sebastian Schuster, Ranjay Krishna, Angel Chang, Li~Fei-Fei, and Christopher~D
  Manning. 2015.
\newblock Generating semantically precise scene graphs from textual
  descriptions for improved image retrieval.
\newblock In {\em Proceedings of the fourth workshop on vision and language\/}.
  volume~2.

\bibitem[{Shi et~al.(2017)Shi, Huang, and Lee}]{DBLP:conf/emnlp/ShiHL17}
Tianze Shi, Liang Huang, and Lillian Lee. 2017.
\newblock Fast(er) exact decoding and global training for transition-based
  dependency parsing via a minimal feature set.
\newblock In {\em {EMNLP}\/}. Association for Computational Linguistics, pages
  12--23.

\bibitem[{Teney et~al.(2016)Teney, Liu, and van~den
  Hengel}]{DBLP:journals/corr/TeneyLH16}
Damien Teney, Lingqiao Liu, and Anton van~den Hengel. 2016.
\newblock Graph-structured representations for visual question answering.
\newblock {\em CoRR\/} abs/1609.05600.

\bibitem[{Vendrov et~al.(2015)Vendrov, Kiros, Fidler, and
  Urtasun}]{DBLP:journals/corr/VendrovKFU15}
Ivan Vendrov, Ryan Kiros, Sanja Fidler, and Raquel Urtasun. 2015.
\newblock Order-embeddings of images and language.
\newblock {\em CoRR\/} abs/1511.06361.

\bibitem[{Wang et~al.(2015)Wang, Xue, and Pradhan}]{DBLP:conf/naacl/WangXP15}
Chuan Wang, Nianwen Xue, and Sameer Pradhan. 2015.
\newblock A transition-based algorithm for {AMR} parsing.
\newblock In {\em {HLT-NAACL}\/}. The Association for Computational
  Linguistics, pages 366--375.

\bibitem[{Werling et~al.(2015)Werling, Angeli, and
  Manning}]{DBLP:conf/acl/WerlingAM15}
Keenon Werling, Gabor Angeli, and Christopher~D. Manning. 2015.
\newblock Robust subgraph generation improves abstract meaning representation
  parsing.
\newblock In {\em {ACL} {(1)}\/}. The Association for Computer Linguistics,
  pages 982--991.

\bibitem[{Xu et~al.(2017)Xu, Zhu, Choy, and Fei{-}Fei}]{DBLP:conf/cvpr/XuZCF17}
Danfei Xu, Yuke Zhu, Christopher~B. Choy, and Li~Fei{-}Fei. 2017.
\newblock Scene graph generation by iterative message passing.
\newblock In {\em {CVPR}\/}. {IEEE} Computer Society, pages 3097--3106.

\bibitem[{Zhu et~al.(2016)Zhu, Groth, Bernstein, and
  Fei{-}Fei}]{DBLP:conf/cvpr/ZhuGBF16}
Yuke Zhu, Oliver Groth, Michael~S. Bernstein, and Li~Fei{-}Fei. 2016.
\newblock Visual7w: Grounded question answering in images.
\newblock In {\em {CVPR}\/}. {IEEE} Computer Society, pages 4995--5004.

\end{thebibliography}
